\documentclass{article}

\usepackage{times}
\usepackage{latexsym}
\usepackage{microtype}
\usepackage{graphicx}
\usepackage{subfigure}
\usepackage{graphicx}
\usepackage{amsmath}
\usepackage{amssymb}
\usepackage{xargs}
\usepackage{placeins}
\usepackage{booktabs} 
\usepackage{array,multirow}

\usepackage{hyperref}

\usepackage{tabularx}
\usepackage[subtle]{savetrees}
\usepackage{enumitem}
\usepackage{url}

\graphicspath{ {./} }

 \usepackage[accepted]{icml2018}

\icmltitlerunning{InclusiveFaceNet: Improving Face Attribute Detection with Race and Gender Diversity}

\begin{document}

\twocolumn[
\icmltitle{InclusiveFaceNet: \\
Improving Face Attribute Detection with Race and Gender Diversity}



\icmlsetsymbol{equal}{*}

\begin{icmlauthorlist}
\icmlauthor{Hee Jung Ryu}{goog}
\icmlauthor{Hartwig Adam}{equal,goog}
\icmlauthor{Margaret Mitchell}{equal,goog}
\end{icmlauthorlist}

\icmlaffiliation{goog}{Google, USA}

\icmlcorrespondingauthor{Hee Jung Ryu}{imiheej@gmail.com}
\icmlcorrespondingauthor{Margaret Mitchell}{mmitchellai@google.com}

\icmlkeywords{Machine Learning, ML Fairness, Fairness, Algorithmic Fairness, Computer Vision, ICML, FATML}

\vskip 0.3in
]



\printAffiliationsAndNotice{\icmlEqualContribution} 

\begin{abstract}

We demonstrate an approach to face attribute detection that retains or improves attribute detection accuracy across gender and race subgroups by learning demographic information prior to learning the attribute detection task. The system, which we call InclusiveFaceNet, detects face attributes by transferring race and gender representations learned from a held-out dataset of public race and gender identities. Leveraging learned demographic representations while withholding demographic inference from the downstream face attribute detection task preserves potential users' demographic privacy while resulting in some of the best reported numbers to date on attribute detection in the Faces of the World and CelebA datasets.\vspace{-.7em}

\end{abstract}

\section{Introduction}

Detecting and recognizing different face attributes has become an increasingly feasible machine learning task due to the rise in deep learning techniques and large people-focused datasets \cite{Niu_2016_CVPR,liuetal2015celeba}. However, this rise has come at a cost:  As public technology incorporates modern computer vision techniques, we are observing a troubling gap between how well some demographics are recognized compared with others \cite{whitehouse2016bigdata}.
For example \cite{gendershades}, the darker and the more feminine the face is the worse commercial gender classifiers performed.

\begin{figure}[h!]
    \centering
    \includegraphics[scale=0.2]{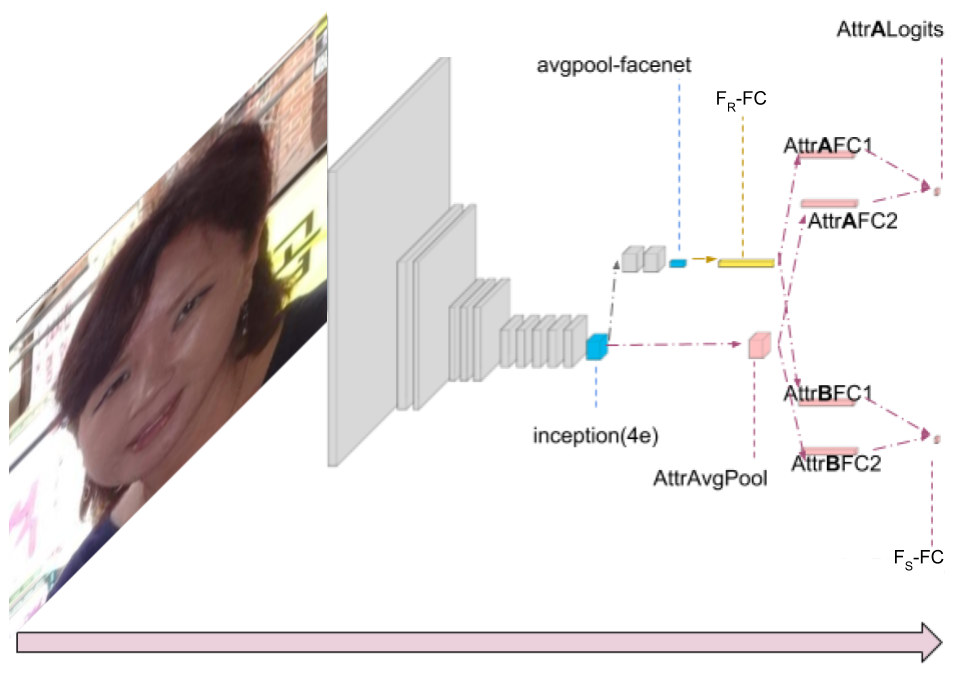}\vspace{-1.25em}
    \caption{{\bf End-to-end Architecture of InclusiveFaceNet.} Layers from the face recognition model are transferred to our diversity classifier and multi-head face attribute detector InclusiveFaceNet. Yellow denotes the decoupled race layer. Pink denotes layers trained from scratch for face attribute detection.}
    \vspace{-1.2em}
    \label{fig:smile_end2end}
\end{figure}

In this paper, we examine the problem of unequal performance across different race and gender subgroups in the context of face attribute detection. We focus on {\bf inclusion}, which we use in this paper to mean improving accuracy across minority subgroups; and {\bf demographic privacy}, where our approach to embracing coarse-grained demographic characteristics that affect the visual characteristics of face attributes does not require demographic inference on users of the downstream face attribute task. Inclusion is related to but not identical to notions of fairness, where the goal is to obtain equal performance across all subgroups (even if that means lower accuracy). Our approach is similar in spirit to the recently introduced proposal for {\it decoupled classifiers} \cite{DworkEtAl18}, where the learning of sensitive attributes can be separated from a downstream task in order to maximize both fairness and accuracy. As in \cite{DworkEtAl18}, we use transfer learning.  While \cite{DworkEtAl18} uses its sub-type, domain adaptation, we use its different sub-type, \textit{task transfer learning}. In our setting, both gender and race are learned as independent, coupled classifiers, and are not inferred at run time.



InclusiveFaceNet matches or improves over a  baseline without transfer learning, defining a new state-of-the-art performance for face attribute detection across gender boundaries in both the CelebA \cite{liuetal2015celeba} and FotW \cite{escaleraetal2017chalearn} datasets.\vspace{-1em}

\section{Ethical Considerations}\label{sec:ethical}
{\bf Intent.} The intent of this work is to demonstrate the utility
of reasoning about demographics -- in this case, race and
gender -- in an ethical and legal setting.

{\bf Race and Gender categories.} Race and gender categories are imperfect for many reasons. For example, there is no gold-standard for `race' categories, and it is unclear how many race and gender categories should be stipulated (or whether they should be treated as discrete categories at all). Throughout our experiments and in this paper, we obscure race and gender names in order to avoid introducing unconscious societal bias associated with existing labels, and anticipate more nuanced treatment of the sensitive categories in future work.

Race and gender identities in this work are learned from a held-out set of public identifications, and the learned representations are then transferred for learning the face attribute models. This permits face attribute detection to leverage race and gender representation without ever having to uniquely predict these characteristics on another individual.\vspace{-.5em}

\section{Background \& Related Work}\label{sec:related_work}

Research in computer vision has included work on issues that have direct social impact, such as security and privacy, however, research on the related issue of diversity and inclusion in vision is surprisingly lacking (but see, e.g., \cite{buolamwini2016ted}). There are several other strands of research background relevant to the work in this paper: face attribute detection and machine learning fairness.

{\bf Face Attribute Detection.} The ChaLearn  ``Looking at People'' challenge from 2016 \cite{escaleraetal2017chalearn} provides the Faces of the World (FotW) dataset, which annotates gender and the presence of smiling on faces. \cite{zhangetal2016gendersmile} won first place in this challenge, utilizing multi-task learning (MTL) and fine-tuning on top of a model trained for face recognition \cite{parkhi2015face}. \cite{ranjan2017smilingbestfotw} later published an out-performing result for the same task on FotW utilizing MTL and transfer learning from a face recognition model \cite{sankar2017face}.
\cite{walkandlearn2016} defined state-of-the-art performance on CelebA face attributes \cite{liuetal2015celeba}, employing both MTL and transfer-learning from a face recognition model based on FaceNet \cite{facenet2016}. This work is similar to the approach taken here, but includes additional processing and data, e.g., geo-location.




\cite{escaleraetal2017chalearn} points out the importance of demographic diversity of
the source material in a vision dataset, and postulates that analysis made on
skewed datasets cannot be representative enough for benchmarking progress.  The authors provide Faces of the World (FotW), collected with the aim to achieve a uniform distribution across two genders and four ethnic groups, and we use this dataset in the current paper (but note that we still find demographic skew).



{\bf Fairness in Machine Learning.}  \cite{DworkEtAl2012} demonstrates the importance of ``Fairness through Awareness'', understanding sensitive characteristics like gender and race in order to build demographically inclusive models.  Proposals for fairness have included {\it parity}, such as Demographic Parity \cite{DworkEtAl2012,Hardtetal2016equality,beutel2017}, 
and equality measures such as that of \cite{Hardtetal2016equality} Equality of Odds, which requires equal false negative rates and false positive rates across subgroups.

Very recent work from \cite{DworkEtAl18} discusses theory related to the approach here, where the authors demonstrate how to define the most accurate classifier as the classifier that simultaneously minimizes false positives and false negatives. They introduce the notion of {\it decoupled classifiers}, in which a separate classifier is trained on each sensitive subgroup, and positive instances of one sensitive subgroup are paired against negative instances across all subgroups. The approach in this paper is similar in that sensitive subgroups are learned separately and transferred, but the classifiers in this work are coupled and learned independently. We anticipate that a decoupled approach would lead to further gains in future work. \vspace{-.5em}


\section{Learning Diversity}\label{sec:learning_diversity} 

\paragraph{Hypothesis.}  At the core of this work lies the idea that faces look different across different races and genders \cite{fuetal2014race}.  If such a relationship holds, then face attribute detection could be improved by learning about race and gender characteristics.\vspace{-.5em}

\paragraph{Evaluation Metrics.}  In addition to overall accuracy, we also provide accuracy per demographic subgroup to examine the effect of our experiments on each race and gender subgroup. Inspired by recent work in machine learning fairness (see Section \ref{sec:related_work}), we also evaluate with a metric that averages the false positive rate and false negative rate, Average False Rate (AFR), which is robust to label and subgroup imbalances in the test data.\vspace{-.3em}



\subsection{Twofold Transfer Learning}

Transfer learning is useful in the case where there is a feature space $\mathcal{X}$ and a marginal distribution over the feature space $P(X_S)$, where $X_S=x_1,x_2 ... x_n \in \mathcal{X}$, as well as a marginal distribution over the related feature space $P(X_D)$, where $X_D \in \mathcal{X}$.  Here, $X_S$ and $X_D$ are completely disjoint sets, respectively representing face images for face attribute detection and face images for race classification.  Formally, given:\vspace{-1em}
\begin{itemize}
\item A face recognition domain $F_X$\vspace{-.5em}
\item A face domain $F_R$ for the task of race classification $T_R$ with labels $Y_R$ and images $X_R$\vspace{-.5em}
\item The target face domain $F_S$ for face attribute detection $T_S$ with labels $Y_S$ and images $X_S$\vspace{-.5em}
\end{itemize}

We optimize for $T_R$ by learning $P(Y_R|X_R)$ via transfer learning from a network trained in the face recognition space $F_X$ with the pre-trained network frozen, and then freeze the learned representations from $F_X$ and $F_R$ in the target domain $F_S$ when learning $P(Y_S|X_S)$. We refer to the use of transfer learning twice in this way as \textbf{twofold transfer learning}.  The second fold of the transfer learning permits the model to leverage demographic representations without inferring demographic characteristics when deployed.\vspace{-.5em}

\subsection{Learning Demographic Diversity} \label{sec:sampling}



Given face crops similar to those from Picasa, we train a FaceNet model \cite{facenet2016} and extract features from the layer called \textit{avgpool}. We add
fully connected layers and then fine-tune the model for the race and gender classification tasks. Race and gender are learned from a held-out dataset with a uniform distribution across race and gender intersections. Our gender model's performance on the publicly available FotW and CelebA are 93.87\% and 99\%, respectively, setting the new state-of-the-art gender classification accuracy.\vspace{-1em} 


\begin{table}[t]
\begin{center}
\small
\resizebox{\columnwidth}{!}{
\begin{tabular}{@{}l@{\hspace{.1em}}c@{\hspace{.3em}}rc@{\hspace{.3em}}c@{\hspace{.3em}}c@{\hspace{.3em}}c@{\hspace{.3em}}cc@{\hspace{.3em}}c@{\hspace{.3em}}l@{}}
& & \multirow{2}{*}{{\bf Total}} & \multicolumn{5}{c}{Estimated Race} & \multicolumn{3}{c}{Gender} \\
& & & {\footnotesize {\bf S1}} & {\footnotesize {\bf S2}} & {\footnotesize {\bf S3}} & {\footnotesize {\bf S4}} & {\footnotesize {\bf Other}} & {\footnotesize {\bf G1}} & {\footnotesize {\bf G2}} & {\footnotesize {\bf Other}} \\
\toprule
\multirow{2}{*}{\rotatebox[origin=c]{90}{\textsc{{FotW}}}}
  &\multicolumn{1}{c}{\small {\bf Train}}
& 6171 & 8\% & 48\% & 11\% & 15\% & 19\% & 54\% & 48\% & 2\% \\
& & & & & \vspace{-1em} & \\
&{\small {\bf Valid}}
& 3086 & 8\% & 53\% & 11\% & 17\% & 11\% & 44\% & 55\% & 1\% \vspace{.25em}\\\midrule

\multirow{3}{*}{\rotatebox[origin=c]{90}{\textsc{{CelebA}}}}
&
{\small {\bf Train}}
& 162687 & 7\% & 75\% & 7\% & 8\% & 4\% & 58\% & 42\% & -\\
&{\small {\bf  Valid}}
& 19863 & 7\% & 76\% & 6\% & 6\% & 6\% & 58\% & 42\% & - \\
&{\small {\bf Test}}
&  19955 & 9\% & 69\% & 9\% & 9\% & 6\% & 61\% & 39\% & -\\
\bottomrule
\end{tabular}
}
\vspace{-1em}\end{center}
\caption{{\bf Estimated Race \& Groundtruth Gender Distribution across Datasets.}
Duplicated images in CelebA are counted only once in this table.
\vspace{-1.5em}
}\label{tab:distro}
\end{table}

\paragraph{Race Classification.}  We explore four race subgroups for which we were able to scrape over 100,000 web images of famous identities from \cite{wikidata2017,rothe2016imdbwiki,usc2017raceclassifier}, and train  
to 98\% or greater AUC across all subgroups.  We do not name the race subgroups here, as our intention is to not to label individuals, but to understand whether learning demographic categories can provide further gains across demographic categories (see Section \ref{sec:ethical}).\vspace{-1em}

\paragraph{InclusiveFaceNet Architecture.} 
We train a multihead face attribute detector which we call InclusiveFaceNet.  Figure \ref{fig:smile_end2end} provides a high-level overview of the architecture used for learning to detect face attributes, denoted with AttrA-* and AttrB-*.\vspace{-1em}




\begin{table*}
\begin{center}
\resizebox{2.05\columnwidth}{!}{
\begin{tabular}{@{}l|l@{\hspace{.5em}}l@{\hspace{.5em}}|l@{\hspace{.5em}}l@{\hspace{.5em}}|l@{\hspace{.5em}}l@{\hspace{.5em}}|l@{\hspace{.5em}}l@{\hspace{.5em}}|l@{\hspace{.5em}}l@{\hspace{.5em}}||l@{\hspace{.5em}}l@{\hspace{.5em}}|l@{\hspace{.5em}}l@{\hspace{.5em}}||l@{\hspace{.5em}}l@{}}
\toprule
Train: FotW & \multicolumn{2}{c|}{\bf Subgroup 1} & \multicolumn{2}{c|}{\bf Subgroup 2} & \multicolumn{2}{c|}{\bf Subgroup 3} & \multicolumn{2}{c|}{\bf Subgroup 4} & \multicolumn{2}{c||}{\bf Other} & \multicolumn{2}{c|}{\bf Gender 1} & \multicolumn{2}{c||}{\bf Gender 2} & \multicolumn{2}{c}{\bf Total} \\
~Eval: FotW & {\small {\sc Acc.\%}} & {\small {\sc AFR\%}}
& {\small {\sc Acc.\%}} & {\small {\sc AFR\%}}
& {\small {\sc Acc.\%}} & {\small {\sc AFR\%}}
& {\small {\sc Acc.\%}} & {\small {\sc AFR\%}}
& {\small {\sc Acc.\%}} & {\small {\sc AFR\%}}
& {\small {\sc Acc.\%}} & {\small {\sc AFR\%}}
& {\small {\sc Acc.\%}} & {\small {\sc AFR\%}}
& {\small {\sc Acc.\%}} & {\small {\sc mAFR\% r/g}}
\\\midrule
\textbf{\textsc{smiling}} &
88.5 & 12.2 &
90.9 &  10.7 &
90.0 &  12.2 &
94.3 &  6.2 &
88.2 & 18.4 & 90.2 & 10.6 & 91.0 & 12.5 &
90.57 & 12.0 / 11.6\\
\textbf{\textsc{~~+g}} &
88.5 & {\bf 12.1} &
91.2 & 10.1 &
{\bf 90.9} & {\bf 10.9} &
94.1 & 6.1 &
88.2 & 18.4 & 90.3 & 10.4 & {\bf 91.3} & {\bf 11.6} &
90.76 & {\bf 11.5} / 11.0\\
\textbf{\textsc{~~+r}} &
88.5 & {\bf 12.1} &
{\bf 91.6} & {\bf 9.7} &
90.6 & 11.3 &
{\bf 94.5} & {\bf 5.9} &
88.2 & 18.4 & {\bf 90.7} & {\bf 10.0}& {\bf 91.3} & 11.8 &
{\bf 90.96} & {\bf 11.5} / {\bf 10.9}\\
\midrule
{\bf + 1FC(2048)} &
88.1 & 12.5 &
91.3 & 10.2 &
90.3 & 11.8 &
{\bf 94.5} & {\bf 5.9} &
88.2 & 18.7 & 90.6 & 10.2 & 91.1 & 12.3 &
90.80 & 11.8 / 11.2 \\
\bottomrule
\end{tabular}
}
\vspace{-1em}\end{center}
\caption{Accuracy ({\sc Acc.}), Average False Rate ({\sc AFR}), and mean AFR ({\sc mAFR}) by race and gender subgroup for smiling detection with proposed systems, on FotW. +G denotes transferred representations from the gender head of the diversity classifier, +R from the race head of the diversity classifier.  The last row serves as a baseline: A model with the same number of parameters as the +G, +R, but without the transferred representations. We see improvements across minority subgroups, as well as the plurality subgroups, for both race and gender. The transferred knowledge does not impact the race `Other' category, which is a racially diverse set of people whose representative racial features our diversity classifier is not trained to capture and extract. We exclude the handful of unlabeled FotW gender instances in reporting the Gender {\sc mAFR}.\vspace{-1.5em}}\label{tab:fairnessanalysisfotw}
\end{table*}


\section{Face Attribute Detection Datasets}\label{sec:smile_dataset}\vspace{-.3em}

{\bf Faces of the World (FotW).} FotW\cite{escaleraetal2017chalearn} is originally introduced by
the ChaLearn Looking at People\footnote{\url{http://chalearnlap.cvc.uab.es/challenge/13/track/20/description/}} Smile and Gender Challenge \cite{escaleraetal2017chalearn} with a goal of demographic diversity.  In the dataset, each image displays a single face labeled with smile, and some with gender. See Table \ref{tab:distro} for details. We estimate Subgroup 2 to make up half the data, with the rest as minorities.\vspace{-.3em}



{\bf Large-scale CelebFaces Attributes (CelebA).} CelebA \cite{liuetal2015celeba} has each image labeled with presence of 40 face attributes (see Table \ref{tab:40attrs}), including smiling/not-smiling, and male/not-male. We estimate Subgroup 2 to be the majority, with relatively little data for other groups (Table \ref{tab:distro}).\vspace{-1em}

\section{Experiments}\vspace{-.3em}

\subsection{Analyzing Inclusion}\vspace{-.3em}

Table \ref{tab:distro} details the racial inclusion in FotW and CelebA.  For FotW, the race model estimates a high variance among the subgroups, with the plurality of the data for Subgroup 2.  Subgroup 1 is the smallest of the subgroups, with only 523 training data instances.  
For CelebA, there is less variance across races, with Subgroup 2 the majority.\vspace{-.5em}

\begin{table}
\begin{center}
\small
\resizebox{1.0\columnwidth}{!}{
\begin{tabular}{@{}l|l @{\hspace{.3em}} l @{\hspace{.3em}} l @{\hspace{.3em}} l @{\hspace{.3em}} |l @{\hspace{.3em}} l @{\hspace{.3em}} l @{\hspace{.3em}} l @{\hspace{.3em}} |l @{\hspace{.3em}} l @{\hspace{.3em}} l @{\hspace{.3em}} l@{}}
\hline
 &
\multicolumn{4}{c|}{\bf Big Lips} &
\multicolumn{4}{c|}{\bf Big Nose} &
\multicolumn{4}{c}{\bf Young}
\\
&
\multicolumn{2}{c}{Gender 1} & \multicolumn{2}{c|}{Gender 2} &
\multicolumn{2}{c}{Gender 1} & \multicolumn{2}{c|}{Gender 2} &
\multicolumn{2}{c}{Gender 1} & \multicolumn{2}{c}{Gender 2} \\
 & {\small {\sc Acc.}} & {\small {\sc AFR}}
& {\small {\sc Acc.}} & {\small {\sc AFR}}
 & {\small {\sc Acc.}} & {\small {\sc AFR}}
 & {\small {\sc Acc.}} & {\small {\sc AFR}}
 & {\small {\sc Acc.}} & {\small {\sc AFR}}
 & {\small {\sc Acc.}} & {\small {\sc AFR}}
\\\midrule
\textbf{\textsc{van.}} &
63.7 & 45.0 & 82.7 & 35.8 &
90.4 & 36.7 & 77.2 & 24.9 &
90.7 & 24.2 & 86.3 & 15.2
\\
\textbf{\textsc{~~~~+r}} &
\textbf{64.6} & \textbf{43.1} & \textbf{82.8} & \textbf{34.9} &
90.4 & \textbf{36.5} & \textbf{77.8}& \textbf{24.6} &
\textbf{90.8} & \textbf{22.8}  & \textbf{86.8} &  \textbf{14.4}
\\\bottomrule
\end{tabular}
}
\vspace{-1em} \caption{\textbf{Example Improved Face Attributes in CelebA Evaluated by Gender.}  Examples of some of the most improved attributes, evaluated by gender. Accuracy (Acc.) and average false rate (AFR) are in percentage (\%).  \textbf{\textsc{van.}}~refers to \textit{vanilla}. +R refers to our model with race input.  We see an improvement up to .9\% absolute.\vspace{-1.5em}}\label{tab:40attrs-by-gender}
\end{center}
\end{table}

\begin{table}
\begin{center}
\small
\resizebox{1.0\columnwidth}{!}{
\begin{tabular}{@{}l|l @{\hspace{.3em}} l @{\hspace{.3em}} l @{\hspace{.3em}} l @{\hspace{.3em}} |l @{\hspace{.3em}} l @{\hspace{.3em}} l @{\hspace{.3em}} l @{\hspace{.3em}} |l @{\hspace{.3em}} l @{\hspace{.3em}} l @{\hspace{.3em}} l@{}}
\hline
 &
\multicolumn{4}{c|}{\bf Oval Face} &
\multicolumn{4}{c|}{\bf Brown Hair} &
\multicolumn{4}{c}{\bf Wavy Hair}
\\
&
\multicolumn{2}{c}{Gender 1} & \multicolumn{2}{c|}{Gender 2} &
\multicolumn{2}{c}{Gender 1} & \multicolumn{2}{c|}{Gender 2} &
\multicolumn{2}{c}{Gender 1} & \multicolumn{2}{c}{Gender 2} \\
 & {\small {\sc Acc.}} & {\small {\sc AFR}}
& {\small {\sc Acc.}} & {\small {\sc AFR}}
 & {\small {\sc Acc.}} & {\small {\sc AFR}}
 & {\small {\sc Acc.}} & {\small {\sc AFR}}
 & {\small {\sc Acc.}} & {\small {\sc AFR}}
 & {\small {\sc Acc.}} & {\small {\sc AFR}}
\\\midrule
\textbf{\textsc{van.}} &
\textbf{74.6} & \textbf{33.4} & \textbf{77.2} & \textbf{44.2} &
\textbf{86.8} & \textbf{18.8} & \textbf{90.1} & \textbf{22.9} &
\textbf{80.6} & \textbf{19.4} & 88.7 & \textbf{32.2}
\\
\textbf{\textsc{~~~~+r}} &
74.2 & 34.3 & 77.1 & 45.5 &
86.5 & 18.2 & 89.9 & 22.7 &
80.4 & 19.7 & 88.7 & 32.6
\\\bottomrule
\end{tabular}
}
\vspace{-1em}\caption{\textbf{Example Face Attributes with Loss in CelebA Evaluated by Gender.}  Examples of some of the attributes with the most loss, evaluated by gender. Accuracy (Acc.) and average false rate (AFR) are in percentage (\%).  VAN. refers to \textit{vanilla}. +R refers to our model with race input.  We see a loss up to .4\% absolute.\vspace{-1.2em}}\label{tab:40attrs-by-gender-loss}
\end{center}
\end{table}

\begin{table}[h!]
\begin{center}
\small
\begin{tabular}{@{}l@{\hspace{.5em}}|l@{\hspace{.5em}}l@{\hspace{.5em}}l@{\hspace{.5em}}l@{\hspace{.5em}}l@{\hspace{.5em}}l@{\hspace{.5em}}l@{\hspace{.5em}}l@{\hspace{.5em}}l@{\hspace{.5em}}l@{\hspace{.5em}}l@{\hspace{.5em}}l@{\hspace{.5em}}l@{\hspace{.5em}}l@{\hspace{.5em}}l@{\hspace{.5em}}l@{\hspace{.5em}}l@{\hspace{.5em}}l@{}}
\hline
 &
\rotatebox{90}{\shortstack[l]{\bf Black\\\bf Hair}} &
\rotatebox{90}{\bf Blurry} &
\rotatebox{90}{\shortstack[l]{\bf Bushy\\\bf Eyebrows}} &
\rotatebox{90}{\bf Chubby} &
\rotatebox{90}{\bf Goatee}&
\rotatebox{90}{\bf Male} &
\rotatebox{90}{\shortstack[l]{\bf Narrow\\\bf Eyes}} &
\rotatebox{90}{\shortstack[l]{\bf Pale\\\bf Skin}} &
\rotatebox{90}{\shortstack[l]{\bf Wear.\\\bf Necklace}} & \rotatebox{90}{\bf Young} \\
\hline
\textbf{\textsc{la}} & 
88 & 84 & 90 & 91 & 95 & 98 & 81 & 91 & 
71 & 87 \\
\textbf{\textsc{wl}} & 
84 & 91 & 93 & 89 & 92 & 96 & 79 & 85 & 
77 & 86 \\
\textbf{\textsc{van.}} & 
89.7 & 96.1 & 92.7 & 95.8 & 97.3 & 98.8 & 86.9 & 96.7 & 
86.3 & 89.2 \\
\textbf{\textsc{~~~+r}} & 
\textbf{89.8} & \textbf{96.2} &  \textbf{92.8} & \textbf{95.9} & \textbf{97.4}& \textbf{99.0} & \textbf{87.0} & \textbf{96.8} & 
\textbf{86.4} & \textbf{89.3} \\\bottomrule
\end{tabular}
\vspace{-1em}\caption{\textbf{Example Improved Face Attributes in CelebA.}  Transfer learning with race representations matches or improves face attribute detection accuracy for 35 out of 40 attributes, with marginal but consistent gains improving 13 attributes over the baseline model, and setting new state-of-the-art on 10 face attributes (presented here). Prior art compared are LA=LNets+ANet \cite{panda2013} and WL=Walk\&Learn \cite{walkandlearn2016}.\vspace{-2em}}\label{tab:40attrs}
\end{center}
\end{table}

\subsection{Evaluating}\label{sec:evaluating}\vspace{-.3em}



Results on Faces of the World are in Table \ref{tab:fairnessanalysisfotw}.  We achieve an accuracy higher than that of the prior art \cite{ranjan2017smilingbestfotw}, using both race and gender \textsc{smiling+r+g}, and set a new state-of-the-art performance of {\bf 90.96\%} with the model that includes a learned race representation, \textsc{smiling+r}.


On FotW, our smile model \textsc{smiling+r+g} reaches an accuracy 1.62\% (absolute) higher than the 2016 ChaLearn Looking at People Challenge winner \textsc{SIAT\_MMLAB} \cite{zhangetal2016gendersmile}, and 0.13\% (absolute) higher than the published state-of-the-art \textsc{UMIACS} \cite{ranjan2017smilingbestfotw} on the FotW validation dataset. 
Our model outperforms prior arts \cite{deepbe2016,uricar2016,Ehrlich_2016_CVPR_Workshops} with an increased performance gap as large as 11.66\% absolute.  The only point in the evaluation where transferred knowledge {\it hurts} performance is from the addition of gender for Subgroup 4 in FotW, where accuracy is less than that of the baseline \textsc{smiling}. The addition of race improves over the baseline.\vspace{-.2em}

The Faces of the World experiments suggest that learning race subgroups may be particularly useful for inclusive models. We next turn to evaluate the effect of learning race on all 40 face attributes found in the widely used dataset CelebA.  We find that the proposed system sets new record accuracies across 23 face attributes, 
defining the state of the art on the 10 attributes in Table \ref{tab:40attrs}. Race representations also further improve accuracy for individual subgroups: For Gender 1, \textit{bangs, big lips, blurry, chubby, double chin, narrow eyes, pale skin, pointy nose, wearing necklace}, and \textit{young} improve with transfer learned race;  For Gender 2, \textit{attractive, bangs, big lips, big nose,
black hair, bushy eyebrows, goatee, wearing necktie}, and \textit{young} all improve with race representations (see Table  \ref{tab:40attrs-by-gender}).  Improvements are marginal but consistent across the board.\vspace{-.2em}

Among the 10 attributes with accuracy improvements with race input,  \textit{big lips} improved by as large as .9\% absolute for Gender 1.  On the other hand, five out of 40 attributes have accuracy losses with the addition of race input: \textit{oval face, brown hair, heavy makeup, high cheekbones}, and \textit{wavy hair}.  \textit{oval face} has the largest accuracy loss delta of .4\% absolute for Gender 1.  This result coincides with what we expected.  \cite{fuetal2014race} says that there are four racially salient facial features: mouth, eyes, lips, and nose.  And, the top three face attributes in our experiments are closely related to one or more racially salient attributes.  \textit{Big Lips} is closely related to the racially salient attribute \textit{mouth}.  \textit{Big Nose} is closely related to the racially salient attribute \textit{nose}.  \textit{Young} is closely related to the racially salient attribute \textit{eyes} and \textit{mouth} according to \cite{young}.  Meanwhile, the bottom three attributes are not closely related to the racially salient attributes.\vspace{-.5em}


\section{Discussion}\label{sec:discussion}

Building an inclusive model and building a state-of-the-art model can go hand-in-hand. This does not have to come at the expense of predicting characteristics that may be sensitive to predict at run-time.

Transferring race knowledge increases the accuracy in almost all cases in FotW and for 13 of the CelebA attributes. Notably, learning race improves performance for smiling across both genders in FotW.   Transferring gender knowledge similarly leads to accuracy and AFR improvements, although it is not as impactful as race.  In FotW, the addition of gender features increases the accuracy of the gender groups as well as race subgroups 2 and 3, in addition to improving accuracy overall. 




Qualitatively, the addition of race and gender representations seem to improve smiling detection on faces with extreme head poses, black-and-white photos, and more diverse facial expressions such as open mouth.  And, albeit incorrect groundtruth of marking neutral faces as smiling, the addition of race and gender representations seem to correctly categorize those neutral faces as not smiling.  This work establishes the feasibility of using transfer learning with demographics to improve performance across demographic categories, which we refer to as algorithmic {\it inclusion}. Future work could examine the effects of joint training and decoupled classification.\vspace{-1em}

\begin{table}[t]
\begin{center}
\resizebox{1.0\columnwidth}{!}{
\begin{tabular}{l|llllll}
\hline
& \multicolumn{2}{c}{\bf \textsc{s}} & \multicolumn{2}{c}{\bf \textsc{s+g}} & \multicolumn{2}{c}{\bf \textsc{s+r}}
\\
& $p$-$value$ & ${\chi}^2$ & $p$-$value$ & ${\chi}^2$ & $p$-$value$ & ${\chi}^2$
\\
\hline
\small{\bf \textsc{s}} & - & - & 5.7${e}^{-7}$ & 25.0 & 1.0${e}^{-4}$ & 15.1
\\
\small{\bf \textsc{s+g}} & 5.7${e}^{-7}$ & 25.0 & - & - & 3.9${e}^{-2}$* & 2.0*
\\
\small{\bf \textsc{s+r}} & 1.0${e}^{-4}$ & 15.1 & 3.9${e}^{-2}$* & 2.0* & - & -
\\
\hline
\end{tabular}
}
\end{center}
\caption{{\bf McNemar Test.} (*) indicates P-value from an exact binomial test and ${\chi}^2$ from McNemar's test with continuity correction.\vspace{-1.5em}}\label{tab:mcnemar}
\end{table}


\section{Acknowledgements}\label{sec:ack}

Special thanks to Google, Blaise Ag{\"u}era y Arcas and Corinna Cortes for their support and sponsorship.  Thanks to Charina Chou and Google legal team for policy and legal advice. Thanks to Marco Andreetto, Timnit Gebru, Sergio Guadarrama, Moritz Hardt, Gautam Kamath, David Karam, Shrikanth Narayanan, Caroline Pantofaru, Florian Schroff, Krishna Somandepalli, Rahul Sukthankar and Jason Yosinski for helpful conversations. Thanks to Ben Hutchinson, Yangsun Lee, Divya Tyam and Andrew Zaldivar for donating their face photos for our research.\vspace{-.5em}


\bibliographystyle{icml2018}

\end{document}